\documentclass[letterpaper, 10 pt, conference]{ieeeconf}  

\IEEEoverridecommandlockouts                              

\usepackage{graphics} 
\usepackage{amsmath} 
\usepackage{amssymb}  
\usepackage{algorithm}
\usepackage{float}
\usepackage{subfig}
\usepackage{algorithmic} 
\usepackage{adjustbox} 
\usepackage[T1]{fontenc}
\usepackage{multicol}
\usepackage{multirow}
\usepackage{dblfloatfix}
\usepackage{array}
\usepackage{subfig}
\usepackage{graphicx}
\usepackage[cmyk]{xcolor}
\usepackage{cite}
\usepackage{placeins}
\usepackage{booktabs}
\usepackage{balance}
\usepackage{caption}
\usepackage{amsmath}
\usepackage{listings}

\usepackage{hyperref}
\usepackage{cleveref}
\hypersetup{hidelinks}

\usepackage{epstopdf}

\title{\LARGE \bf
CalliRewrite: Recovering Handwriting Behaviors from Calligraphy Images without Supervision
}

\author{Yuxuan Luo$^{1,2*}$, Zekun Wu$^{1*}$, Zhouhui Lian$^{1\dagger}$
\thanks{$^{1}$Wangxuan Institute of Computer Technology, Peking University}
\thanks{$^{2}$Yuanpei College, Peking University}
\thanks{$^{*}$Equal contribution}
\thanks{$^{\dagger}$Corresponding author: \tt\small lianzhouhui@pku.edu.cn}
}

\begin{document}

\maketitle
\thispagestyle{empty}
\pagestyle{empty}

\begin{abstract}

Human-like planning skills and dexterous manipulation have long posed challenges in the fields of robotics and artificial intelligence (AI). The task of reinterpreting calligraphy presents a formidable challenge, as it involves the decomposition of strokes and dexterous utensil control. Previous efforts have primarily focused on supervised learning of a single instrument, limiting the performance of robots in the realm of cross-domain text replication. To address these challenges, we propose CalliRewrite: a coarse-to-fine approach for robot arms to discover and recover plausible writing orders from diverse calligraphy images without requiring labeled demonstrations. Our model achieves fine-grained control of various writing utensils. Specifically, an unsupervised image-to-sequence model decomposes a given calligraphy glyph to obtain a coarse stroke sequence. Using an RL algorithm, a simulated brush is fine-tuned to generate stylized trajectories for robotic arm control. Evaluation in simulation and physical robot scenarios reveals that our method successfully replicates unseen fonts and styles while achieving integrity in unknown characters. To access our code and supplementary materials, please visit our project page: \href{https://luoprojectpage.github.io/callirewrite/}{https://luoprojectpage.github.io/callirewrite/}.
\end{abstract}


\section{INTRODUCTION}

Calligraphy robots exemplify technology and arts, advancing artificial intelligence, exploring human-robot interaction, and serving as educational tools. 

This paper aims to address the gaps in limited writing utensil manipulation and stroke-level learning, which previous research has not fully explored. Besides conventional algorithms like developmental algorithm~\cite{wu2021developmental}, clustering~\cite{8665143}, dynamic programming, and Gaussian process~\cite{yang2021hybrid}, some researchers have also investigated deep learning techniques such as GAN~\cite{bidgoli2020artistic,wu2020integration} and LSTM~\cite{chao2018use} for stylizing and controlling pen strokes. However, these approaches heavily rely on laborious annotations and heavily supervised learning, making them less effective in scenarios like understanding and parsing ancient scripts.

On the contrary, we focus on unsupervised low-resource learning, mining knowledge from plain images, and self-discovering dexterous control over various utensils. We draw inspiration from how humans break down tasks and skillfully use tools. Even when humans lack precise knowledge of stroke order, such as when children first practice writing their names or scholars transcribe newly found ancient characters, they demonstrate effective planning strategies. Moreover, a proficient artist can replicate a specific writing style with different instruments, for example, etching brush-written fonts onto a tablet with a knife. These examples highlight humans' crucial skills in task planning and motion control and are critical to robotics research.

This paper proposes CalliRewrite: an unsupervised approach enabling robotic arms to replicate diverse calligraphic glyphs with generalization on manipulating different writing tools. Our method employs a hierarchical structure encompassing a CNN-encoded LSTM model to deduce stroke-level orders, and a reinforcement learning (RL) pipeline to fine-tune the coarse sequences into tool-aware stylized control, controlling the brush agent with soft-actor-critic (SAC) algorithm. Utilizing self-supervised loss functions and formulating the fine-tuning into a constrained sequence optimization, we enable the model to self-discover applicable stroke orders without any annotations and discover dexterous control on different utensils, especially on Chinese paintbrush, which demands superb skill.
\begin{figure}[t]
    \vspace{2.5mm}
    \centering
    \includegraphics[width=\linewidth]{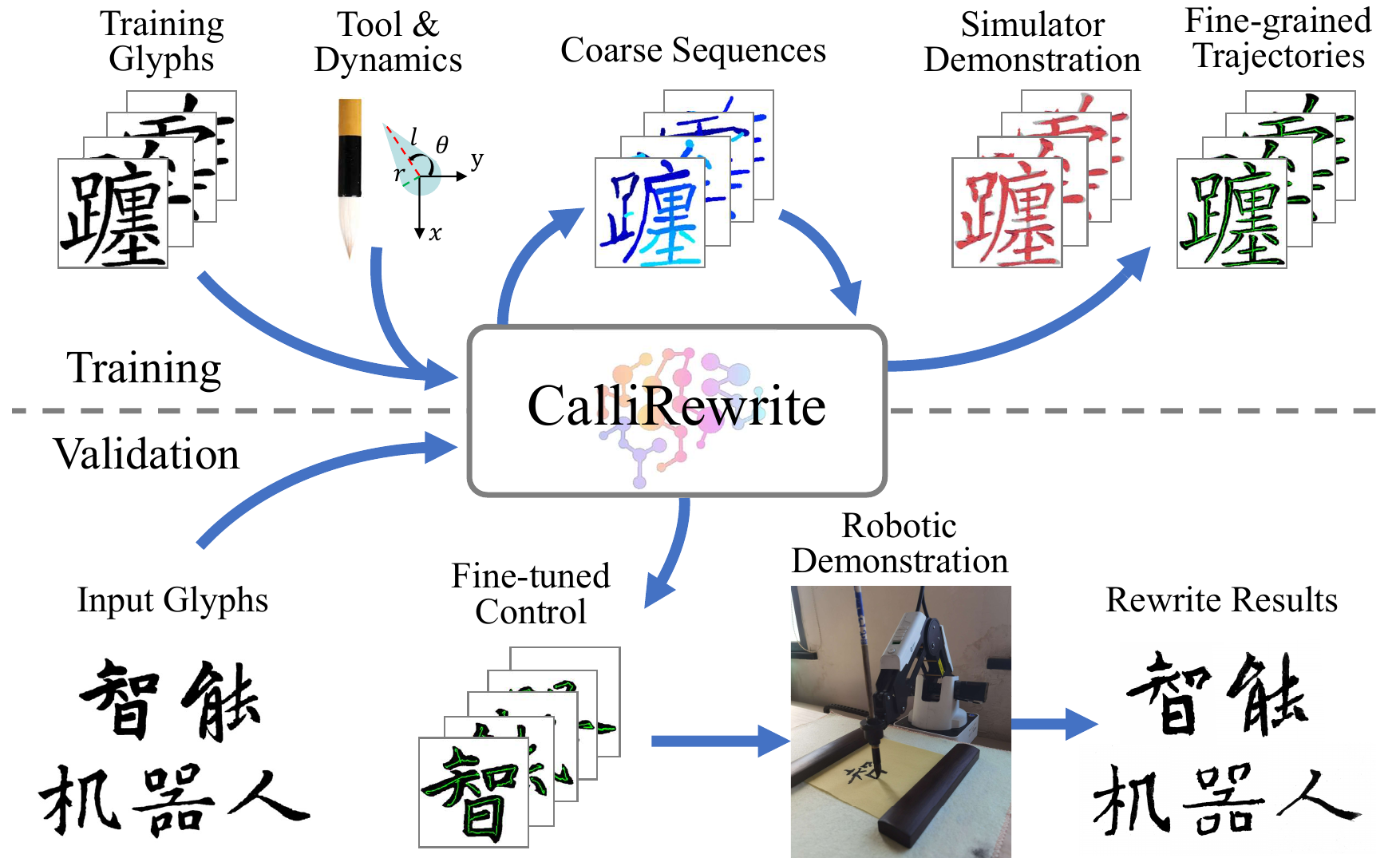}
\caption{Our method, CalliRewrite, enables unsupervised and tool-dependent robot calligraphy. During training, we provide only glyph images and virtual writing utensils. Our two-stage model gradually proposes and fine-tunes applicable writing trajectories conditioned on specific tools. In validation, we apply the parametric fine control on the Dobot robotic arm, which accurately redraws the characters.}
\label{fig: teaser}
\vspace{-6mm}
\end{figure}

We devise a progressive training for low-resource stroke decomposition, pretrained on QuickDraw to learn universal redrawing and better decomposition on 1000 training glyphs. Our evaluation covers diverse calligraphy scripts: Chinese, English, Ancient Egyptian, and Tamil. Our metrics cover stroke number ratio (SNR) and Chamfer Distance, depicting stroke continuity and overall redrawn fidelity respectively. We not only carry out verification on simulations but also rewrite glyphs with a Dobot-Magician robot arm, following the generated control sequence with corresponding utensils.

Experiments reveal that our model surpasses other unsupervised methods in stroke segmentation and performs comparably with supervised models. Additionally, our calligraphy robot can faithfully rewrite unseen fonts or sketches and adeptly control diverse writing implements in unsupervised and low-resource scenarios.
\section{RELATED WORK}

Our work spans glyph comprehension and tool manipulation. Glyph comprehension involves recognizing and decomposing characters, while tool manipulation focuses on modeling utensils and learning dexterous control.

\subsection{Glyph Comprehension}

Glyph comprehension involves character reconstruction and decomposition and is challenging due to human writing's complexity and variability. Traditional models based on expert systems generated glyphs through defined norms and templates~\cite{xu2005automatic}. Deep learning has enabled methods via convolution neural networks (CNNs) and Generative Adversarial Networks (GANs), including "Zi2Zi"~\cite{zi2zi}, "Rewrite"~\cite{rewrite}, PEGAN~\cite{sun2018pyramid}, HAN~\cite{chang2018chinese}, AEGG~\cite{lyu2017auto}, DC-Font~\cite{jiang2017dcfont}. These methods employed image pyramids, hierarchical losses, and refined networks to enhance glyph restoration.

Additionally, Semantic Segmentation~\cite{kim2018semantic} employed CNN and Markovian Random Field for vectorized characters. Berio et al.~\cite{berio2022strokestyles} split strokes by overlapping and intersecting regions. VecFontSDF~\cite{xia2023vecfontsdf} created premium vector fonts using signed distance functions (SDF). DeepVecFont~\cite{wang2021deepvecfont} and DeepVecFont-V2~\cite{wang2023deepvecfont} generated highly editable vector fonts with multimodality. FontTransformer~\cite{liu2023fonttransformer} produced high-resolution Chinese glyph images through few-shot learning.

Some researchers~\cite{tang2019fontrnn,kotani2019teaching,singh2022intelli} adopted trajectories or demonstrations, while others~\cite{yang2021hybrid,8665143} employed manually-split strokes. However, annotating data is laborious, and these approaches have limitations in handling out-of-distribution scenarios like ancient characters. Thus, unsupervised methods are crucial.

Painting and calligraphy share similar origins, but painting allows more freedom in stroke order. Various unsupervised methods used RNN, transformer, and RL to solve the image repainting task, like Sketch-RNN~\cite{ha2017neural}, General Virtual Sketching Framework~\cite{mo2021general}, and Learning to Paint~\cite{huang2019learning}. Schaldenbrandt et al.~\cite{schaldenbrand2021content} extended this to robot artists. Xie et al.~\cite{xie2021dg} proposed deformable generative networks for unsupervised font generation.

Despite these successes, generating well-segmented and realistic writing norms with minimal supervision remains challenging, waiting for improved designs to capture the complexity of human writing.

\subsection{Dexterous Manipulation}

Like dexterous manipulation and automated navigation, calligraphy robots also demand meticulous control. The learned strategies should adjust to different writing implements, such as reproducing a brush-written character using a chisel on a stone tablet. However, in calligraphy robots, the significance of tools is seldom discussed~\cite{wang2023generative}. Traditional trajectory optimization~\cite{clarke1982nonlinear, johnson2005pid} and search algorithms~\cite{lavalle1998rapidly,karaman2011sampling} fell short, while reinforcement learning (RL) offered a paradigm for adaptive tool-based learning~\cite{wu2018towards}. To reduce excessive degrees of freedom (DoF) in exploration~\cite{matheron2020pbcs}, some have attempted a two-stage learning approach, incorporating traditional methods like Bayesian Optimization, Greedy algorithms, or Markovian processes with RL algorithms or curriculum learning~\cite{yang2022learning,zhang2023robot, 9636369}. Others have opted for learn-from-demonstration and inverse RL due to a weariness of reward engineering~\cite{sun2014robot, chao2017robot,xie2023end}.

To autonomously discover tool-specific control techniques, we adopt a coarse-to-fine strategy. Our method applies unsupervised glyph comprehension to initiate stroke guidance and RL trajectory refinement on diverse writing instruments. 

We refer to writing tools as agents. While physical models can faithfully simulate deformations and dynamics~\cite{wong2000virtual,chu2002efficient,xu2003advanced}, they are computationally complex. Therefore, we simplify it by modeling only the geometries and transformations of the contact surface of writing utensils on the canvas~\cite{wang2020robot,lo2006brush}.

\begin{figure} [t!]  
    \vspace{2.5mm}
	\centering
	\subfloat[Coarse Sequence Extraction]{
		\includegraphics[scale=0.275]{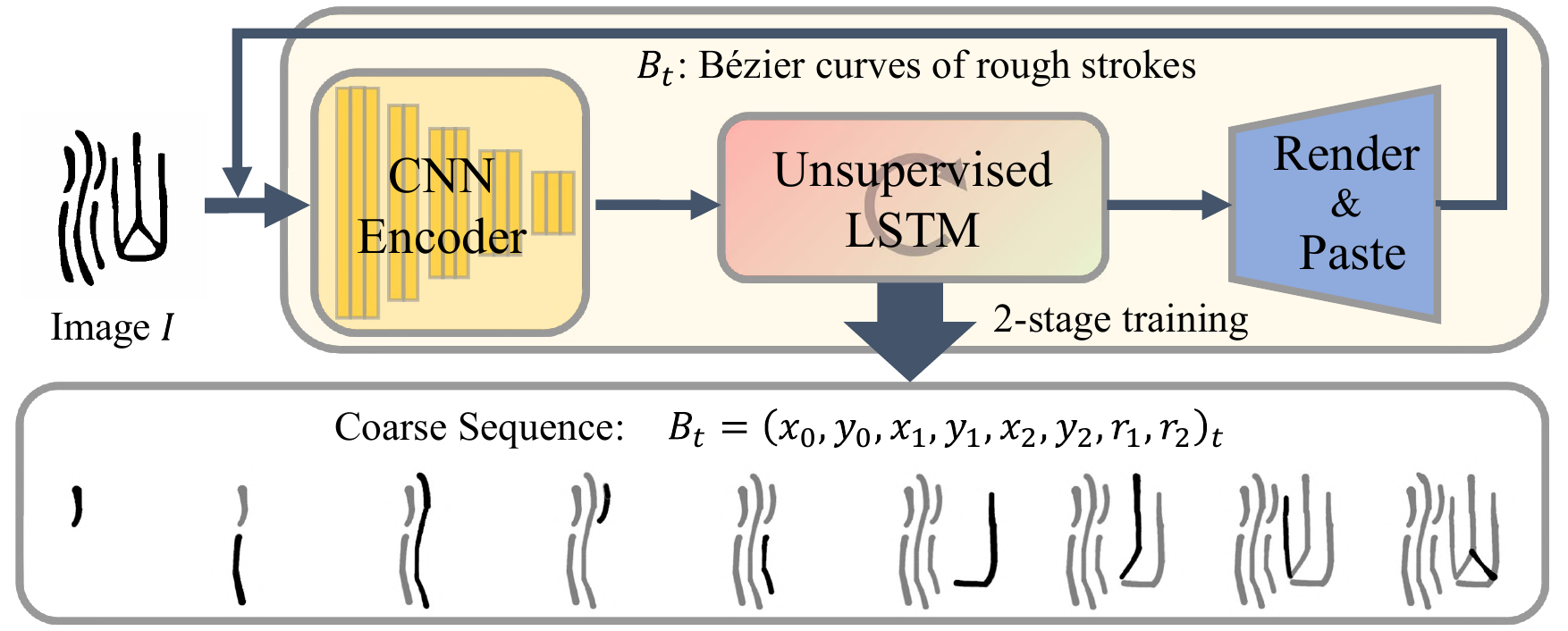}}
    \vspace{-4mm}
    \subfloat[Tool-aware Fine-tuning]{
		\includegraphics[scale=0.275]{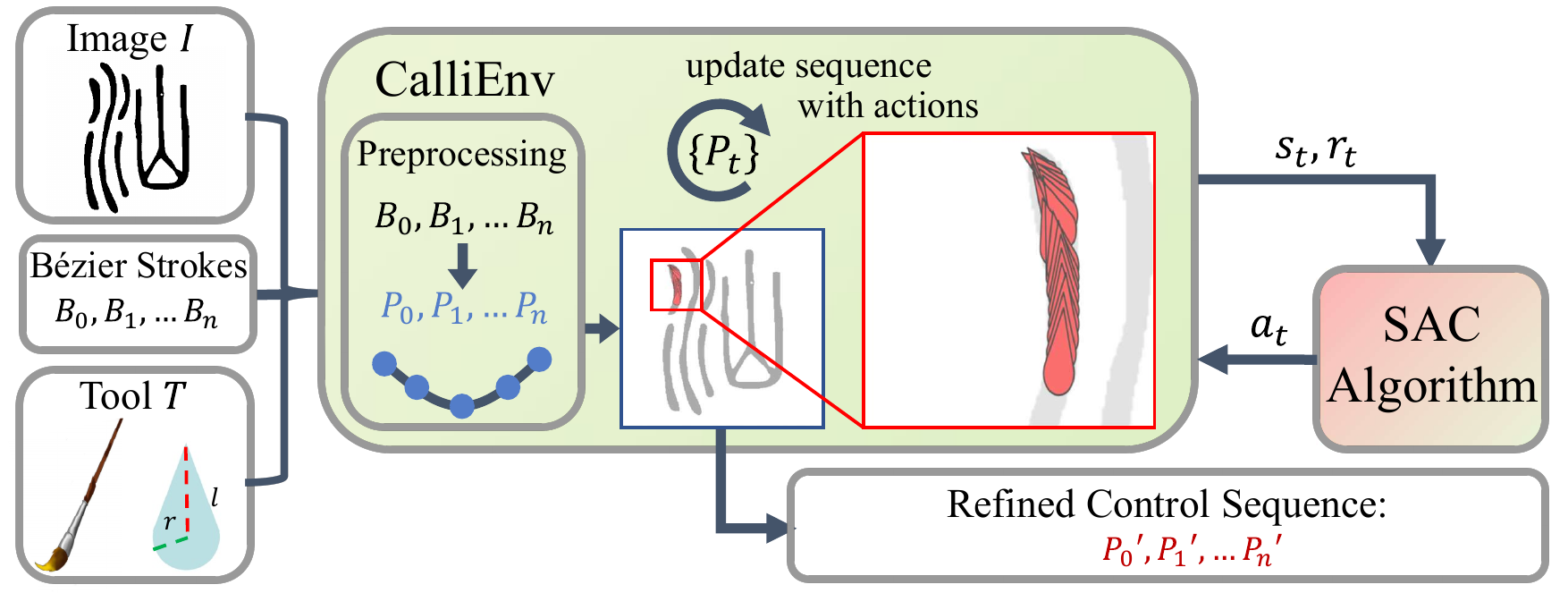}}
	\vspace{-1.5mm}	
	\caption{Method overview: (a) Generating coarse writing trajectories using unsupervised LSTM with tailored loss for human-like glyph decomposition. (b) Reinforcement learning with utensil properties for fine-tuning dexterous control.}
	\label{fig: model} 
\vspace{-5mm}
\end{figure}

\begin{table*}[!htp]
\centering
\vspace{2.5mm}
\caption{Notations}
\vspace{-6.3mm}
\label{tab:reference table}
\begin{multicols}{4}
\begin{tabular}{l p{0.47\linewidth}}
\hline
Notation & Description \\
\hline
$I'$ & redrawn glyph \\
$\gamma$ & discount factor \\
$\mathcal{C}$ & cosine similarity \\
$T$ & total time step \\
$p_t$ & pen state \\
$w_i$ & stroke width \\
\hline
\end{tabular}
\columnbreak
\hspace{-19mm}
\begin{tabular}{l p{0.55\linewidth}}
\hline
Notation & Description \\
\hline

$t$ & current time step \\
$I$ & input glyph image \\
$\rho_t$ & trajectory curvature\\
$\phi$ & perceptual model \\
$S_t$ & 9-dim state space \\
$h_t$ & writing progress \\

\hline
\end{tabular}
\columnbreak
\hspace{-26mm}
\begin{tabular}{l p{0.7\linewidth}}
\hline
Notation & Description \\
\hline
$o_{\text{tool}}$ & occupied utensil pixels\\
$o_{\text{in}}$ & pixels inside glyph\\
$o_{\text{out}}$ & pixels outside glyph\\
IoU & intersection over union \\
$\alpha$ & entropy temperature\\

$\tau$ & soft update coefficient\\

\hline
\end{tabular}
\columnbreak
\hspace{-19mm}
\begin{tabular}{l p{0.59\linewidth}}
\hline
Notation & Description \\
\hline
$\vec{a}_t = (\delta_x, \delta_y, (\theta))_t$ & action space \\
$\mathcal{O}_i = (x_i, y_i)$ & control points \\
$(r, l, \theta)_t$ & utensil geometry \\

$\vec{v}_t = (v_x, v_y)_t$ & stroke direction\\
$B_t$ & quadratic Bézier\\
$\{\mathbf{P}_i\}$ & discrete points on $B_t$ \\

\hline
\end{tabular}
\end{multicols}
\vspace{-2mm}
\vspace{-5mm}
\vspace{-0.5mm}
\end{table*}
\vspace{-1mm}
\section{METHOD}
\vspace{-1mm}
Figure~\ref{fig: model} outlines our two-stage approach: Coarse Sequence Extraction and Tool-aware Fine-tuning. Given the calligraphy glyph $I$, a CNN-LSTM model segments initial strokes into quadratic Bézier curves $\{B_t\}$. We employ a fusion of self-supervised objectives and a progressive training strategy to enhance writing order and integrity with limited training data.

In the second phase, we formulate the task into a constrained reinforcement learning problem, implement various utensils, and tailor the environment with well-crafted rewards. Utilizing the SAC model as a controller, we leverage coarse sequences to curtail ineffective exploration.

The notations adopted in this paper are listed in Table~\ref{tab:reference table}.
\subsection{Extracting Coarse Sequences}
\vspace{-1mm}
\label{coarse sequence extraction}
\subsubsection{Model Design}
\label{model}
Mo et al.~\cite{mo2021general} introduced a framework for generating vector line art but neglected to preserve writing order due to simple loss terms. Our approach retains the foundational architecture while introducing refined objectives to achieve more continuous and feasible decomposition results comparable to human behaviors.

Specifically, we adopt a CNN-LSTM backbone and borrow the dynamic window generation and the differentiable cropping-pasting component. In time step $t$, current canvas with glyph $I$ are convolved to infer a quadratic Bézier stroke $B_t = (p, x_0, y_0, x_1, y_1, x_2, y_2, w_0, w_1)_t$, where $p$ indicates the pen state for drawing ($p=1$) or lifting ($p=0$) the utensil; $\mathcal{O}_i, i\in\{1,2,3\}$ denotes the control points and $w_0, w_1$ represent the initial and final widths, respectively. Drawing from Gestalt theory~\cite{kohler1967gestalt} and physical constraints, we formulate a set of unsupervised losses trained alongside a progressive approach.
\vspace{2mm}
\subsubsection{Unsupervised Losses}
\label{self_supervise_loss}
\mbox{}\\
\textbf{Perceptual loss.}
\label{perceptual loss}
We utilize the VGG-16-based perceptual loss~\cite{johnson2016perceptual} to ensure redrawn similarity. It compares the similarity between the rendered output $\hat{y}$ and the target glyph $y$. The $j$th perceptual loss is
\begin{small}
\vspace{-1mm}
\begin{equation}
    \mathcal{L}_{\text {perc}}^j=\frac{1}{D_j \times H_j \times W_j}\left\|\phi_j(\hat{y})-\phi_j(y)\right\|_1.
\end{equation}
\vspace{-4mm}
\end{small}

To prevent excessive loss fluctuations, we normalize each loss value for each layer with $t$ and compute the total loss at layers \verb|relu1_2|, \verb|relu2_2|, \verb|relu3_3| and \verb|relu5_1| ($J=\{12, 22, 33, 51\}$) of the VGG-16 model:
\vspace{-2mm}
\begin{equation}
    \mathcal{L}_{\text {perc }}=\sum_{j \in J} \mathcal{L}_{\text {perc-norm }}^j = \sum_{j \in J}\frac{\mathcal{L}_{\text {perc }}^j}{\mathcal{L}_{\text {perc-mean }}^j}.
    \vspace{-1mm}
\end{equation}
\textbf{Regularization loss.}
\label{regularization loss}
We introduce the stroke regularization term to reduce redundant strokes by penalizing lifting actions ($p_t=0$):
\vspace{-1.6mm}
\begin{equation}
\vspace{-2.2mm}
    \mathcal{L}_{\text {reg }}=\frac{1}{T} \sum_{t=1}^T (1-p_t).
\end{equation}
\textbf{Smoothness loss.}
\label{smoothness loss}
Decomposing overlapping strokes and intersections is pivotal in glyph recognition. Rooted in the Gestalt theory, the continuity principle underscores our inclination to perceive connected and unbroken elements. To achieve this, we employ the cosine similarity $\mathcal{C}$ between $\overrightarrow{\mathcal{O}_1\mathcal{O}_0}$ and $\overrightarrow{\mathcal{O}_1\mathcal{O}_2}$ to yield generating smoother strokes:
\vspace{-1.5mm}
\begin{equation}
    \mathcal{L}_{\text {smo}}=\sum_{t=1}^T p_t\cdot \frac{1+\mathcal{C}(\overrightarrow{\mathcal{O}_1\mathcal{O}_0}, \overrightarrow{\mathcal{O}_1\mathcal{O}_2})}{2}.
    \vspace{-0.5mm}
\end{equation}
To mitigate false penalization of real-exist curves, we incorporate a scaling factor of 0.5. Consequently, while decomposing a curvature into multiple strokes does reduce $\mathcal{L}_{\text{smo}}$, the consequential increase in  $\mathcal{L}_{\text{reg}}$ is more pronounced. Overall, this results in a higher cumulative loss.

\textbf{Angle loss.}
\label{angle loss}
Physical constraints in the human arm and muscles make us write with a starting direction from a certain angle range, approximately around +75° and -165°. Therefore, for adjacent generated sequences $B_{t-1}$ and $B_t$, if $p_{t-1}=0$ and $p_t=1$, we employ the cosine similarity $\mathcal{C}$ between $\vec{S_t}=(\overrightarrow{\mathcal{O}_0\mathcal{O}_2})_t$ and $\vec{a} = (-\frac{\sqrt{2}}{2}, \frac{\sqrt{2}}{2})$ to constrain the proper writing direction:
\vspace{-2mm}
\begin{equation}
    \mathcal{L}_{\text {ang}}= \frac{1}{T} \sum_{t=1}^T \left\{ 
\begin{array}{ll}
\mathcal{C}(\vec{S_t},\vec{a}) ,&\mathcal{C}(\vec{S_t},\vec{a})\ge 0.5 \\
0,&\mathcal{C}(\vec{S_t},\vec{a})< 0.5.
\end{array}
\right.
\end{equation}
\subsubsection{Progressive Training}
\label{progressive training}
We adopt a progressive training approach. Initially, the model is trained on the QuickDraw dataset using $\mathcal{L}_{\text{phase1}}$. Subsequently, fine-tuning is performed on 1000 glyphs extracted from the KaiTi-GB2312~\cite{kaiti_gb2312} and KanjiVG~\cite{kanjivg} datasets, guided by $\mathcal{L}_{\text{phase2}}$. In the first phase, the model gains fundamental decomposition capabilities. This allows the model in the second phase to improve its decomposition results while efficiently handling variations in calligraphy stroke thickness. Subsequent experiments have confirmed the necessity of this approach. During training, $\lambda_1$ and $\lambda_2$ increase from 0 to 0.5 as training goes on.
\vspace{-2mm}
\begin{equation}
    \mathcal{L}_{\text {phase1}} = \mathcal{L}_{\text {perc }} + \lambda_1 \cdot \mathcal{L}_{\text {reg }},
     \label{stage 1 loss}
\end{equation}
\vspace{-7mm}
\begin{equation}
    \mathcal{L}_{\text {phase2}} = \mathcal{L}_{\text {perc }} + \mathcal{L}_{\text {reg }} + \lambda_2 (0.5 \cdot \mathcal{L}_{\text {smo}} + \mathcal{L}_{\text {ang }}).
     \label{stage 2 loss}
\vspace{-1mm}
\end{equation}
\subsection{Tool-Aware Fine-tuning}
\vspace{-1mm}
\label{tool aware fine-tuning}
\begin{figure} [t!]  
      \centering
      \includegraphics[scale=0.28]{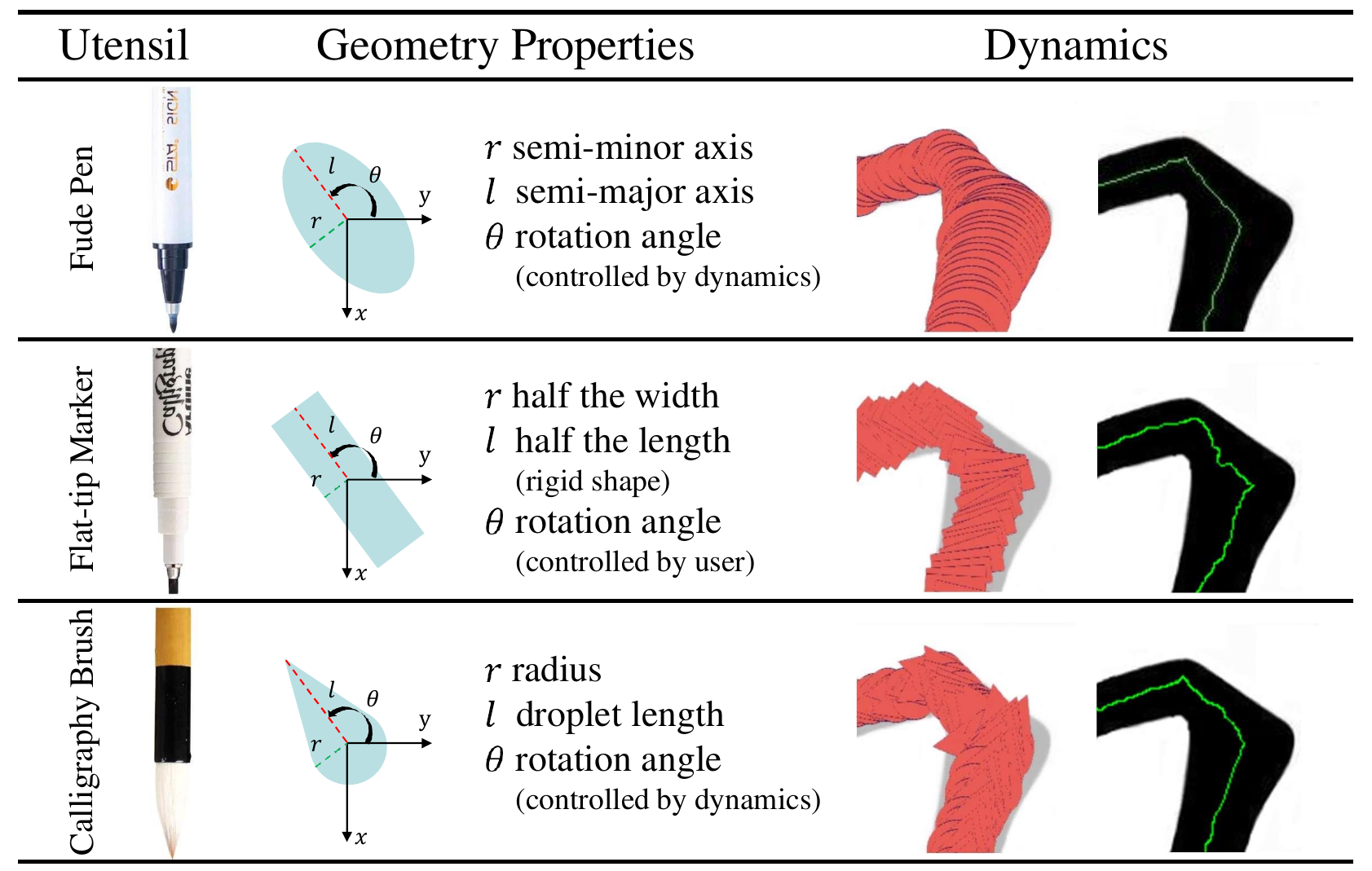}
      \caption{Brief introduction of our modeled writing utensils. More information can be found in the supplemental material.}
      \label{fig: utensil} 
      \vspace{-5mm}
\end{figure}
Previous methods applied segmented strokes to robotic arms without accounting for the impact of writing instruments on control trajectory. We use RL methods to master the plug-and-play utensil models and fine-tune coarse trajectories to improve alignment with real-world robotic scenarios.
\subsubsection {Constrained Sequence Optimization}
\label{sequence}
Calligraphy writing requires thin and precise strokes and challenges sampling effectiveness. Naive exploration leads to out-of-bound collections. Therefore, we focus on the exploration process around the coarse trajectory, avoiding the disruption of ineffective experiences while encouraging the agent to practice diverse utensil control within the boundary. We discretize $B_t$ into $\{\mathbf{P}_i\}_t$: $\mathbf{P}_i = (p, x, y)_i$ and restrain the exploration zone as $\mathbf{P}_t + \vec{a}_t$, where the RL model predicts the action $\vec{a}_t$. 
\subsubsection{Model Design}
\label{algorithm}
We use Soft Actor-Critic (SAC) due to its stability and quick convergence. It adheres to the actor-critic paradigm, featuring two critics to decrease overestimation bias. The objective is to maximize entropy and total reward to encourage exploration and prevent overfitting. The importance of entropy is weighted by $\alpha$ to promote randomness. The configurations about the networks and other hyperparameters are available in Table~\ref{tab:model_configuration}.
\begin{table}[!t]
\vspace{2mm}
\centering
\caption{The SAC Model and Training Configurations}
\vspace{-2mm}
\label{tab:model_configuration}
\begin{tabular}{llll}
\toprule
Actor-Network & \multicolumn{2}{l}{($|s|$, 64, 128, $|a|$)} \\
Critic-Network & \multicolumn{2}{l}{($|s|+|a|$, 128, 128, 128, 1)} \\
\midrule
Parameter & Value & Parameter & Value\\
\midrule
$\alpha$ & 0.95 & Optimizers & Adam \\
$\gamma$ & 0.95 & Replay Buffer & $2^{20}$ \\
$\tau$  & 0.95 & Batch Size & $2^{10}$ \\
policy\_lr & 1e-4 & Training Epochs & 150 \\
q\_lr & 1e-4 & Steps per Epoch & 10000 \\
\bottomrule
\end{tabular}
\vspace{-4.5mm}
\end{table}

\subsubsection{CalliEnv Environment}
\label{environment}
We customize the RL environment with several crucial designs.

\textbf{Plug-and-play utensil models.}
We instantiate writing tools by considering their geometric attributes, encompassing two length measurements ($r$ and $l$), an angular parameter ($\theta$), and their dynamic functions. We select three tools, displayed in Figure~\ref{fig: utensil}: the fude pen, flat-tip marker, and calligraphy brush, covering rigid and flexible forms varying from circular to quadrangular shapes. The parameters and movement dynamics are all approximation results from the real world, which will be discussed in the supplementary.

\textbf{State and action spaces.}
The vector state at timestep $t$, denoted as $S_t$, comprises a 9-dimensional tensor: $S_t = (h, r, l, \theta, \rho, \delta_x, \delta_y, v_x, v_y)_t$, where $h$ denotes the current stroke progress; $r$, $l$, and $\theta$ represents utensil geometry; $\rho$ signals the current curvature; $\delta_x$, $\delta_y$ represent prior actions as offsets; and $v_x$, $v_y$ signify the moving direction within the ongoing stroke.

For most circumstances, the action space defines the displacement coordinates around the point $\mathbf{P}_t$, represented by $\vec{a}_t = (\delta_x, \delta_y)_t$. In the case of tools like a flat-tip marker, the user's wrist orientation, rather than dynamics, controls the angular attribute $\theta$. Thus, we add a new DoF $\theta$ into $\vec{a}_t$.

The RL algorithm controls the agent to move through each $\mathbf{P}_t$ and apply actions, adjusting the scattered points to $\mathbf{P}_t+\vec{a}_t$. The model manipulates the writing utensil iteratively to generate precise control points $\{\mathbf{P}_t'\}$ for writing.
\subsubsection{Reward Design}
\label{reward}
We tailor the rewards for the RL algorithm to learn skillful control and boost convergence.

\textbf{Adaptive shape reward.}
In calligraphy, versatile control is applied at stroke beginnings and ends with fewer variations in the middle. To encourage exploring diverse approaches, we formulate a composite reward, fostering smooth strokes in the middle while imposing penalties for actions that exceed glyph boundaries at the endpoints, seeking control to align with the writing tool:
\vspace{-1mm}
\begin{equation}
\label{eqn:r1}
\mathcal{R}_{\text{ada}} = 
\left\{
\begin{array}{ll}
-\sqrt{\frac{r_{\text{max}}}{r_t}}\cdot \left\| \frac{1}{r_t}-\frac{1}{r_{t-1}}\right\| ,& h_t\in[0.2, 0.8]\\
\frac{o_{\text{in}} - o_{\text{out}}}{o_{\text{tool}}}, & h_t\in[0, 0.2)\cup (0.8,1],\\
\end{array}
\right.
\end{equation} where $o_{\text{tool}}$ represents pixels occupied by the writing tool, $o_{\text{in}}$  represents the overlapping pixels between the tool and the glyph, and $o_{\text{out}}$ covers the tool area outside stroke boundaries.

\textbf{Terminal reward.}
 To ensure redrawn similarity, we apply the Intersection over Union (IoU)~\cite{rezatofighi2019generalized} to compare the redrawn and initial glyph. We multiply the termination reward by a factor of 80 to improve the sensitivity of the SAC algorithm to reward shaping:
\vspace{-2mm}
\begin{equation}
    \mathcal{R}_{\text{fin}} =80\cdot \text{IoU}(I', I).
    \vspace{-2mm}
\end{equation}
In summary, the total reward for a period is:
\vspace{-2mm}
\begin{equation}
    \mathcal{R} = \frac{1}{T}\sum\limits_{t=1}^T \mathcal{R}_{\text{ada}} + \mathcal{R}_{\text{fin }}.
\vspace{-1mm}
\end{equation}

\section{EXPERIMENTS}
\vspace{-1mm}
We quantitatively and qualitatively evaluate our approach to diverse Chinese and English scripts, affirming its versatility. Trained solely on QuickDraw and Kaiti-GB2312, our method successfully rewrites characters in Tamil and Ancient Egyptian using the Dobot Magician robot arm.
\vspace{-2mm}
\subsection{Datasets and Implementation Details}
\vspace{-2mm}
\subsubsection{Training}
Training efficiently with limited data poses challenges. We construct a dataset of 1000 glyphs from KaiTi-GB2312~\cite{kaiti_gb2312} and KanjiVG~\cite{kanjivg}. Employing progressive training, we first train the sequence extraction module on QuickDraw for 75,000 steps and then perform an additional 30,000 steps on our dataset. We use the Adam optimizer with a learning rate of 1e-4, set the batch size to 32, and train the first phase on two RTX-2080 GPUs.

\begin{figure}[t]
    \vspace{1.5mm}
    \begin{minipage}{0.05\textwidth}
    \centering
    {\footnotesize{w/ pt}}
    \end{minipage}
    \begin{minipage}{0.25\textwidth}
             \includegraphics[width=\linewidth]{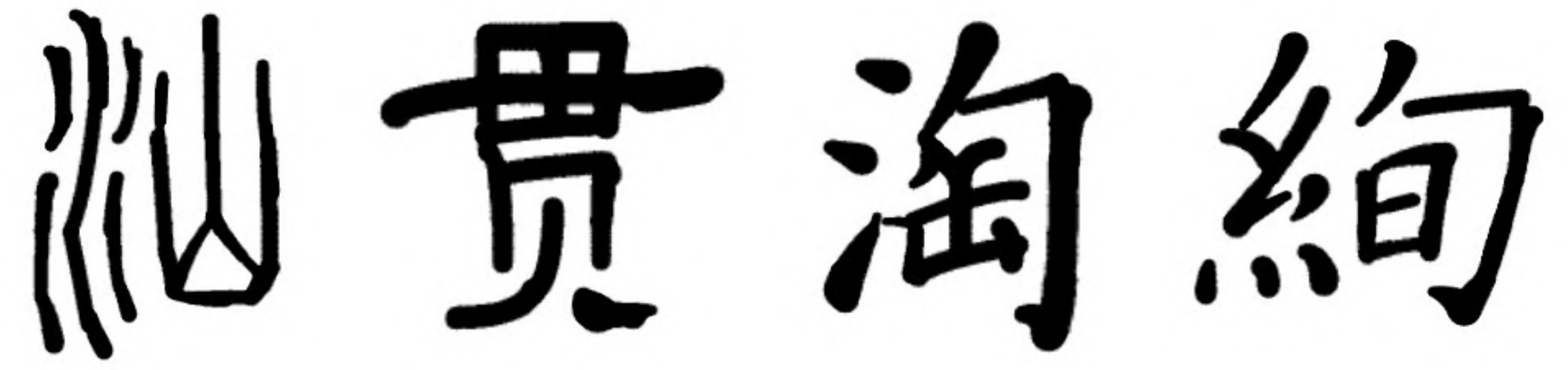}
     \end{minipage}
    \begin{minipage}{0.05\textwidth}
    \centering
    {\footnotesize{w/ ft}}
    \end{minipage}
    \begin{minipage}{0.12\textwidth}
            \includegraphics[width=\linewidth]{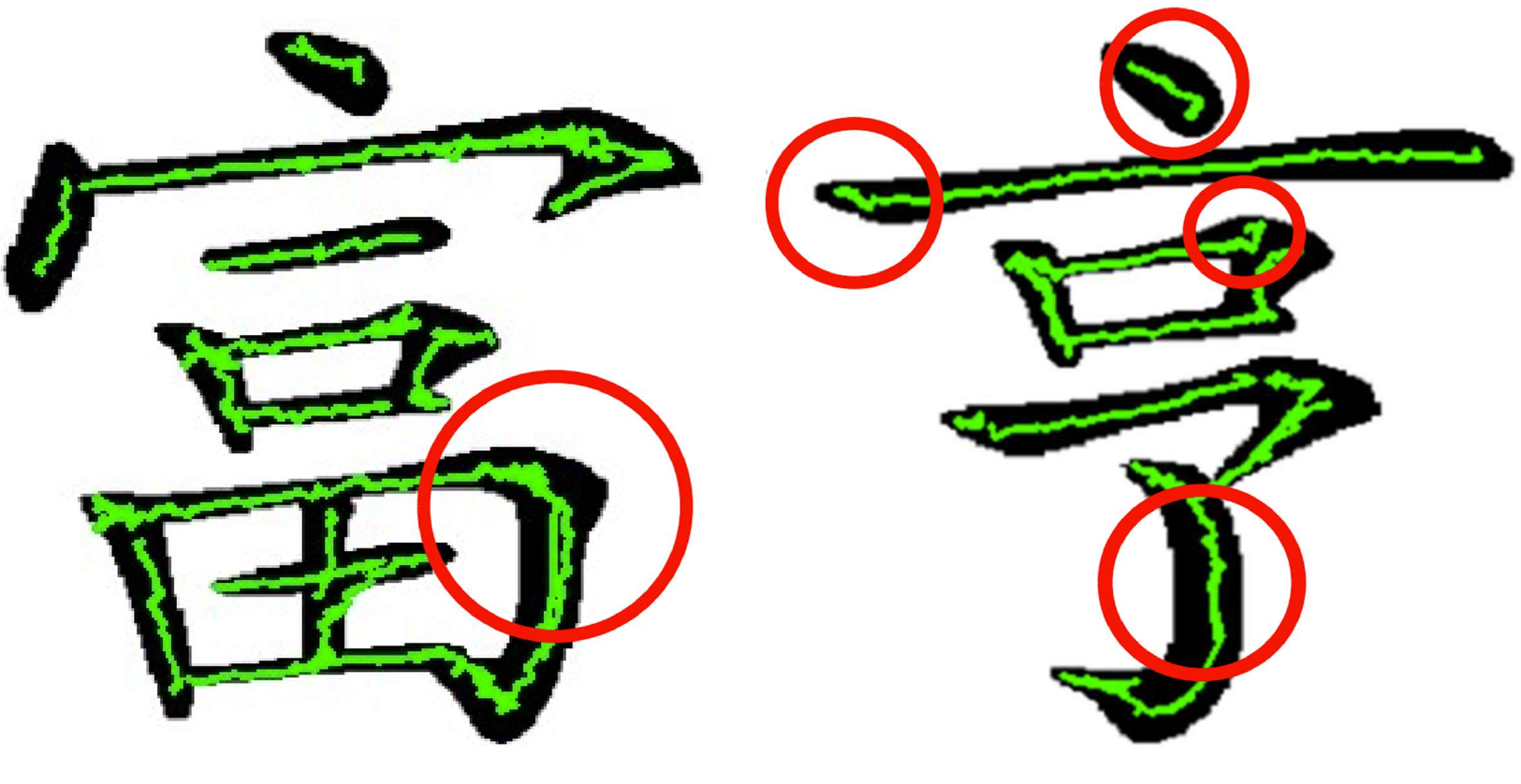}
    \end{minipage}

\caption{(a) Progressive Training ("pt") Comparison. (b) Impact of Fine-Tuning ("ft") on Stroke Connectivity and Tool-Aware Control, including joining disconnected strokes and adding brush-related control at both ends of the strokes.}
\vspace{-4mm}
\label{fig: with_without_ft}
\end{figure}

Figure~\ref{fig: with_without_ft} illustrates results for models trained directly for 75,000 steps on 1000 glyphs and those subjected to progressive training. Progressive training enhances the model's ability to reconstruct glyphs under low-resource conditions.
\begin{figure}[t]
    \vspace{2mm}
    \begin{minipage}{0.05\textwidth}
    \centering
        {\footnotesize{w/o pt}}
    \end{minipage}
    \begin{minipage}{0.25\textwidth}
            \includegraphics[width=\linewidth]{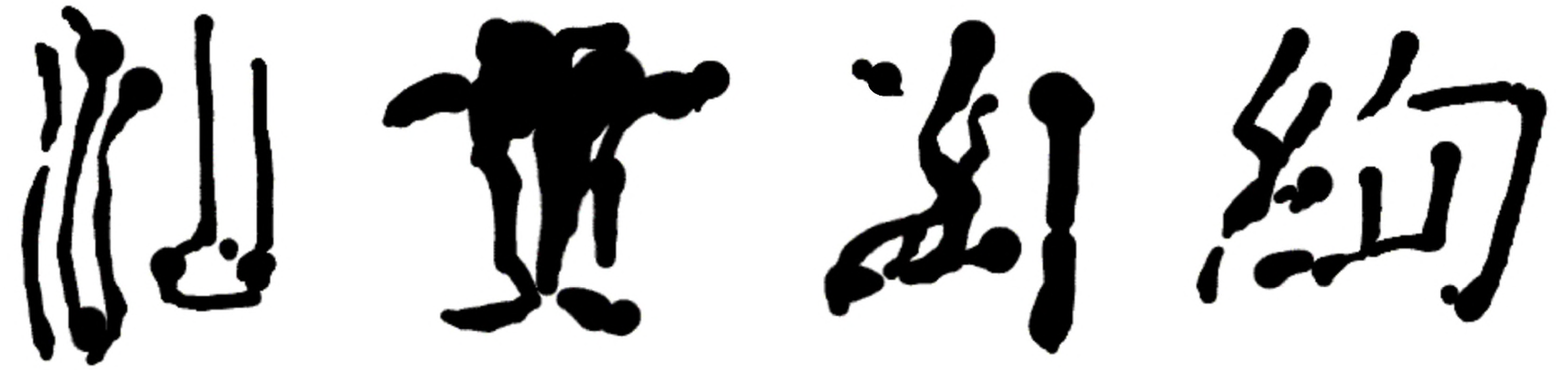}
    \end{minipage}
    \begin{minipage}{0.05\textwidth}
    \centering
    {\footnotesize{w/o ft}}
    \end{minipage}
    \begin{minipage}{0.12\textwidth}
            \includegraphics[width=\linewidth]{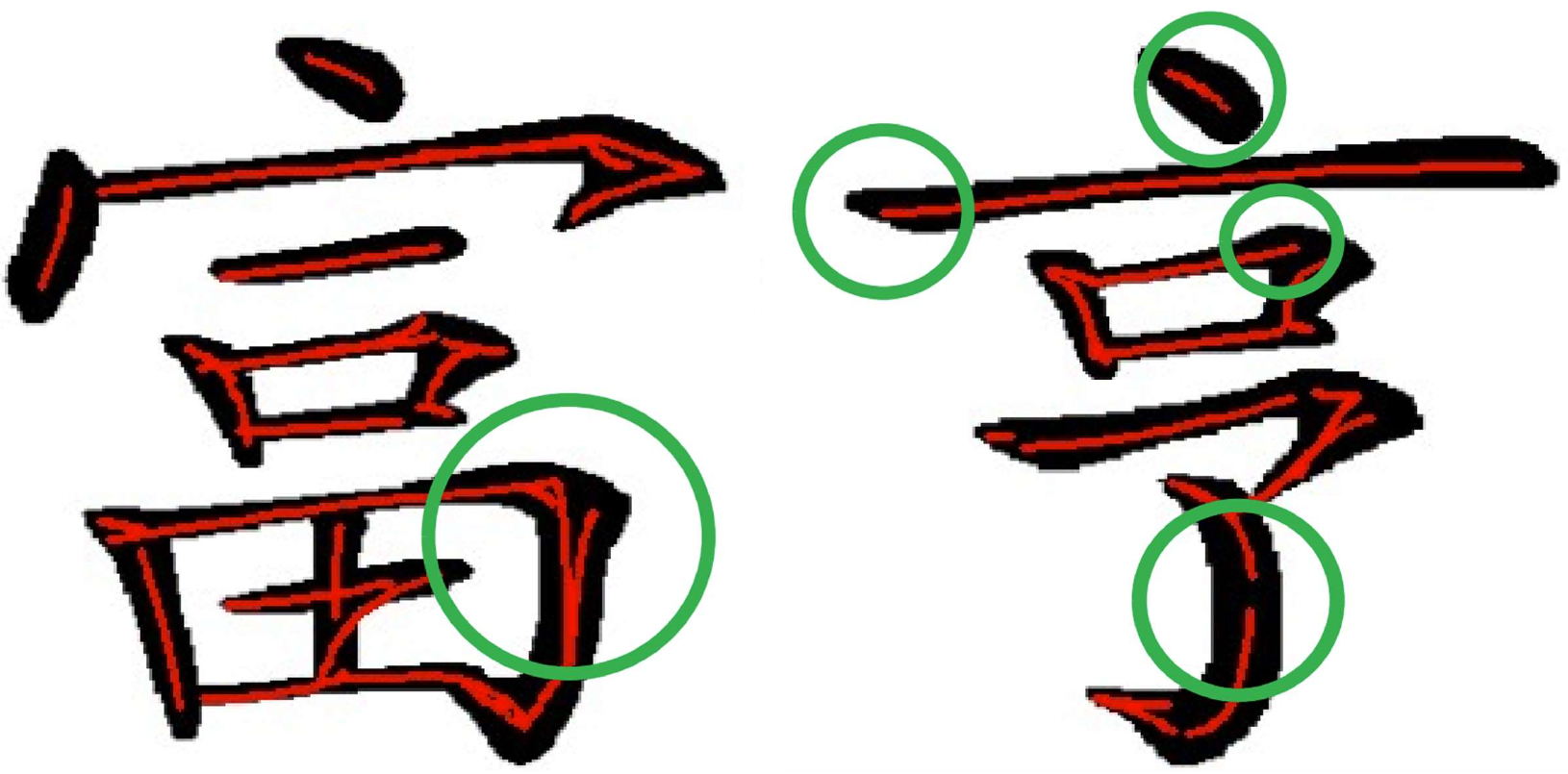}
    \end{minipage}
\end{figure}

We build the CalliEnv environment with OpenAI Gym and Tianshou for modularity and multi-process training (Table~\ref{tab:model_configuration}). The environment utilizes glyph images and decomposition results to generate refined control, with the trained policy for validation.

\subsubsection{Validation}
We evaluate stroke integrity and pixel-wise similarity using Stroke Number Ratio (SNR) and Chamfer Distance (CD). SNR expresses the ratio of partition numbers $N_s$ to the number of ground-truth strokes $N_{gt}$ as $\text{SNR} = \frac{N_s}{N_{gt}}$. A lower SNR value indicates a better stroke continuity. Chamfer Distance, commonly used to compute the similarity between point clouds, is utilized here to calculate the contour similarity between generated and reference glyph images. The definition proposed by Fan et al.~\cite{fan2017point} is adopted to extract and compare the contours.

We use 200 Chinese glyphs in the Kaiti style with annotated stroke numbers for quantitative testing. During robotic experiments, we utilize various languages and scripts like Chinese calligraphy, English, Standard Tamil and ancient Egyptian hieroglyphs. The control sequences are verified through physical tests and renderings on a Dobot-Magician robot arm with writing tools. The results confirm the effectiveness of our method both qualitatively and quantitatively.
\vspace{-6mm}
\subsection{Extracting Coarse Sequences}
\vspace{-1.2mm}
\subsubsection{Low Resource Training}
We examine our model's performance over different training set sizes from 500 to 2000 glyphs. Table~\ref{tab:low resource} shows that our model achieves optimal SNR and CD with just 1000 images while increasing or decreasing the dataset size has minor impacts on the overall performance. This indicates that an LSTM model is enough for the task of learning from small and unsupervised datasets.
\vspace{-4mm}
\subsubsection{Ablation on Unsupervised Losses}
Unsupervised losses $\mathcal{L}_{\text{smo}}$ and $\mathcal{L}_{\text{ang}}$ significantly impact decomposition quality. We perform an ablation study on them to validate our design. Table~\ref{tab:ablation} confirms that combining $\mathcal{L}_{\text{smo}}$ and $\mathcal{L}_{\text{ang}}$ yields the highest effectiveness in terms of both SNR and CD.
\begin{figure}[!t]
\vspace{2mm}
\begin{minipage}{0.42\linewidth}
    \footnotesize
    \captionsetup{justification=centering}
    
    \captionof{table}{Low Resource Training}
    \vspace{-1mm}
    \begin{tabular}{clclc}
        \toprule
        Train Size & SNR ($\downarrow$) & CD ($\downarrow$)\\
        \midrule
        500  & 1.260 & 2.134\\
        \textbf{1000} & \textbf{1.200} & \textbf{1.979}\\
        2000 & 1.204 & 2.153\\
        5000 & 1.238 & 2.135\\
        \bottomrule
    \end{tabular}
    \label{tab:low resource}
\end{minipage}
\begin{minipage}{0.47\linewidth}
    \centering
    \footnotesize
      \captionsetup{justification=centering}
    
    \captionof{table}{Ablation on Unsupervised Losses}
    \vspace{-1mm}
    \begin{adjustbox}{left=7cm}
    \begin{tabular}{clclc}
      \toprule
      Losses & SNR ($\downarrow$) & CD ($\downarrow$)\\
      \midrule
      -- & 5.602 & 2.092\\
      $\mathcal{L}_{\text{smo}}$ & 1.341 & 2.448\\
      $\mathcal{L}_{\text{ang}}$ & 1.376 & 2.134\\
      $\boldsymbol{\mathcal{L}_{\textbf{\text{smo}}}+\mathcal{L}_{\textbf{\text{ang}}}}$ & \textbf{1.199}& \textbf{1.979}\\
      \bottomrule
    \end{tabular}
    \label{tab:ablation}
    \end{adjustbox}
\end{minipage}
\vspace{-3mm}
\end{figure}

\subsubsection{Comparison to Other Methods}
We compare our model to various methods, including "Learning to Paint" (LTP)~\cite{huang2019learning}, "General Virtual Sketching Framework" (GVS)~\cite{mo2021general}, and "VectorNet"~\cite{kim2018semantic}. VectorNet specializes in character segmentation and uses an OverlapNet for stroke intersection prediction, while the former two methods represent stroke-based image repainting models. Both methods are evaluated on 200 Chinese glyphs in the Kaiti style with ground-truth stroke number.

\begin{figure}[t!]
    \centering
    \vspace{2mm}
    \begin{minipage}{0.95\textwidth}
          \subfloat[]{\includegraphics[width=0.33\linewidth]{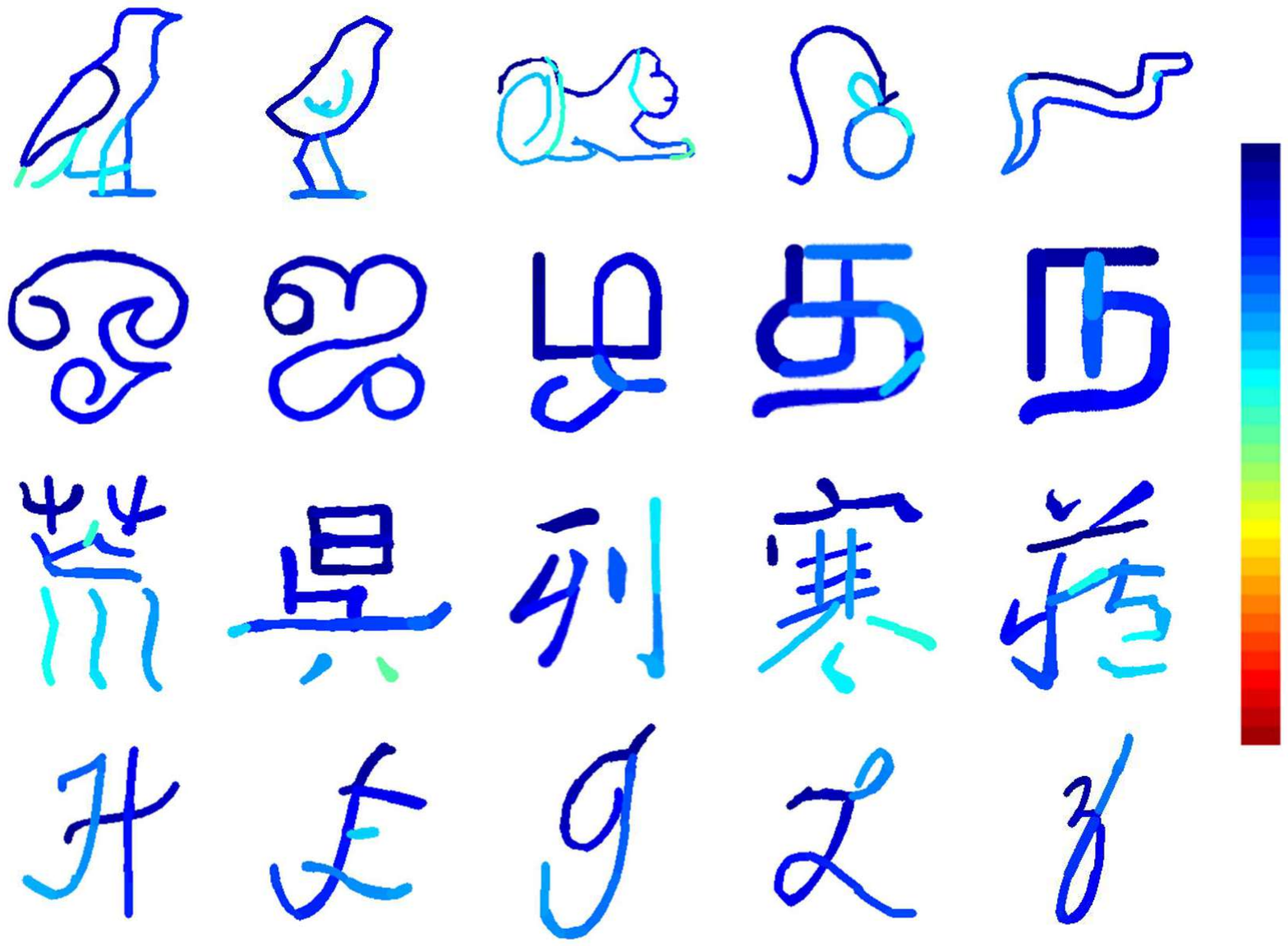}}
          \hspace{1mm}
          \subfloat[]{\includegraphics[width=0.14\linewidth]{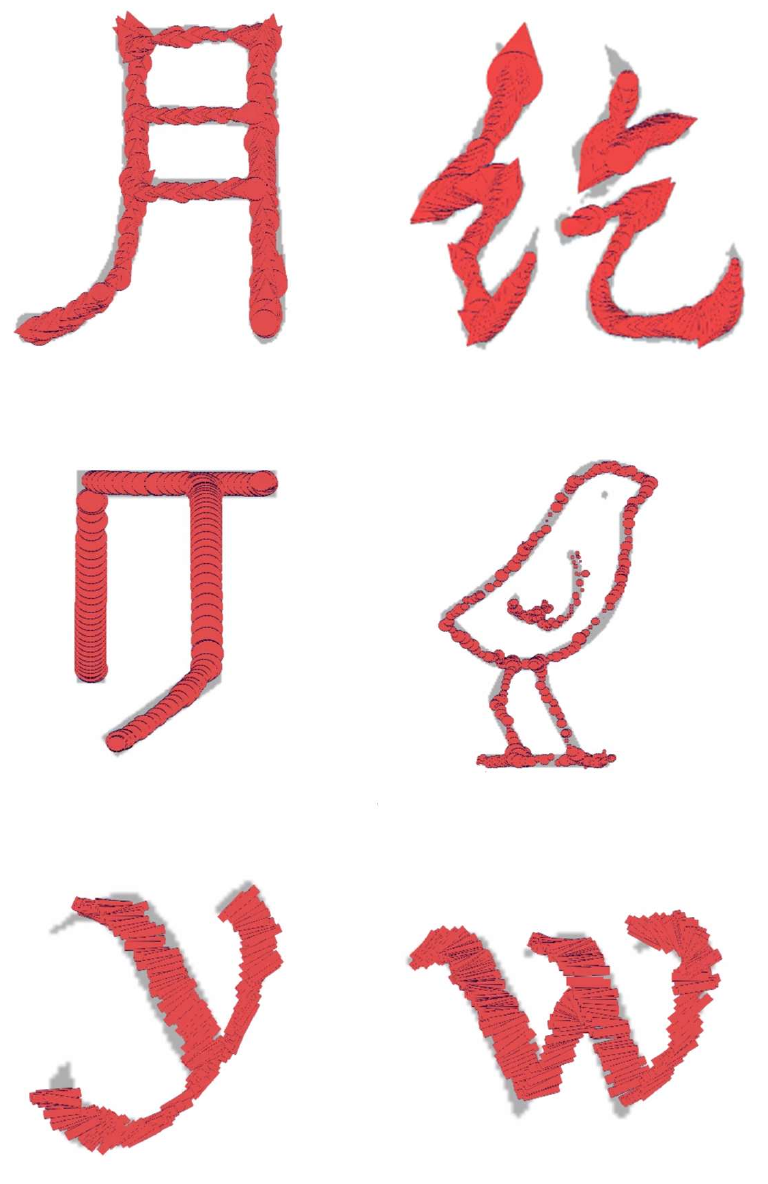}}
    \end{minipage}
\vspace{-2mm}
\caption{(a) Visualization on decomposed coarse sequences. From blue to red: stroke order from the 1st to the 100th.\\ (b) Tool-aware fine-tuning with three writing utensils.}
\label{fig: strokes}
\vspace{-2mm}
\end{figure}

\begin{figure*}[htbp]
    \vspace{2mm}
    \centering
    \begin{minipage}{0.93\textwidth}
          \hspace{-2mm}
          \includegraphics[width=0.3\linewidth]{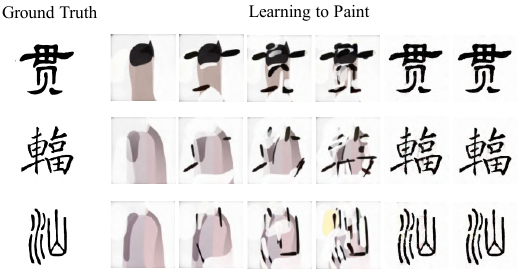}
          \hspace{-0.2mm}
          \includegraphics[width=0.459\linewidth]{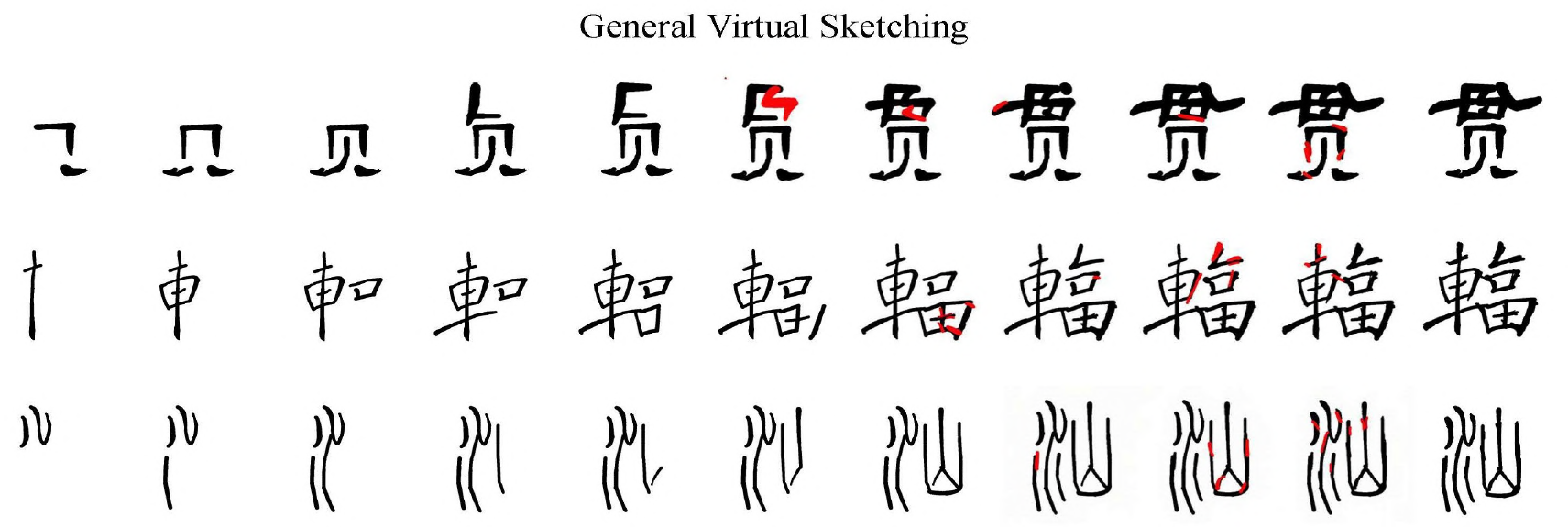}
          \hspace{-1.1mm}
          \includegraphics[width=0.235\linewidth]{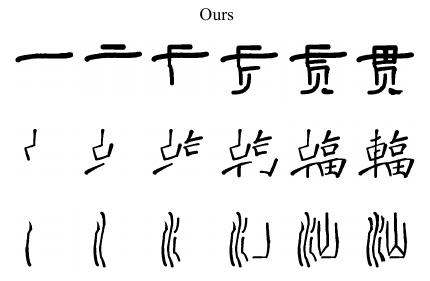}
    \end{minipage}
    \vspace{-1mm}
    \caption{Rendered comparison on diverse unsupervised models: LTP can not generate semantic-aware stroke orders. GVS partially restores ordering but suffers from overlapping and incorrect decomposition (highlighted), affecting stroke integrity. In contrast, our method delivers superior stroke splitting, order, fewer overlaps, and quality similar to GVS in rendering.}
    \label{fig: visual_comparison}
    \vspace{-4mm}
\end{figure*}

In comparison, we limit the maximum strokes of LTP and GVS to 60 and 49 segments. We compute the average SNR and compare Chamfer distances between the reference and the reconstructed image, or vectorized rendered glyph with the size of 256$\times$256 pixels. Table~\ref{tab:comparison} shows that our model outperforms unsupervised methods in SNR and Chamfer distance and is compatible with supervised models, with Figure~\ref{fig: visual_comparison} unfolds our decomposed strokes with better integrity and order while maintaining a high reconstruction quality in rendered images, as also seen in Figure~\ref{fig: strokes}(a).

\begin{table}[t!]
\centering
\footnotesize
\caption{Quantitative Comparison between Methods}
\vspace{-1mm}
\label{tab:comparison}
\begin{tabular}{lclclcl}
\toprule
  & Supervision & SNR ($\downarrow$) & CD ($\downarrow$) \\
\midrule
LTP~\cite{huang2019learning} & \textcolor{blue!80!cyan}{RL} & 5.660 & 6.978 \\
GVS~\cite{mo2021general} & \textcolor{green!70!lime}{Unsupervised} & 5.602 & 2.092 \\
Ours & \textcolor{green!70!lime}{Unsupervised} & \textbf{1.199} & \textbf{1.979} \\
VectorNet~\cite{kim2018semantic} & \textcolor{red}{Supervised} & 1.039 & 1.614 \\
\bottomrule
\end{tabular}
\vspace{-3mm}
\end{table}
\vspace{-2mm}
\subsection{Fine-tuning Dexterous Control}
\vspace{-1mm}
\subsubsection{Devising Mastery over Instruments}
We test various tools in CalliEnv using the SAC algorithm, including calligraphy brushes, fude pens, and flat-tip markers. Different writing tools are used based on the font type: brushes for Chinese, fude pens for rounder English, Ancient Egyptian and Tamil, and flat-tip markers for angular English. Figure~\ref{fig: strokes}(b) displays the visual results of the rewriting process for each tool after 64 training epochs.

 We also explore diverse tools on the same character shape. In Figure~\ref{fig: utensil}, we display the experiments on the Chinese character "Heng Zhe". It's evident that an Ellipse-shaped brush can produce smoother strokes, the marker learns to change direction during turns, and a brush introduces more details at the turning points, including representations of lifting and turning the brush with sharp angles.

\subsubsection{Rectifying Coarse Sequences}
Besides accommodating the coarse sequences to specific instruments, the fine-tuning process can also rectify errors, connecting false breakpoints and stretching the trajectory from a wrong place. As the tool explores the fine-tuned trajectories, it will guide the trajectory toward areas with higher rewards. In Figure~\ref{fig: with_without_ft}(b), we observe that the brush model learns the intricate movements of pen turning at the beginning and end of writing.

\begin{figure*}[htbp]
    \centering
      \begin{minipage}{0.952\textwidth}
       \end{minipage}
    \vspace{-4mm}
    \begin{minipage}{\textwidth}
    \subfloat{
        \begin{minipage}{0.48\textwidth}
                \includegraphics[width=1.0\linewidth]{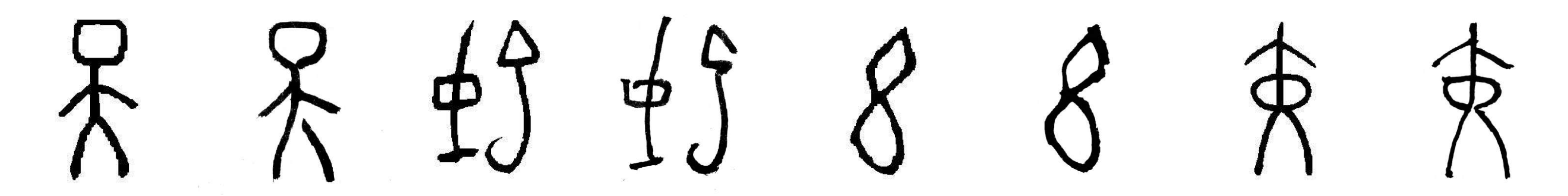}
                \label{fig: ch_result_1}
        \end{minipage}
        \hspace{-3mm}
        \begin{minipage}{0.48\textwidth}
                \includegraphics[width=1.0\linewidth]{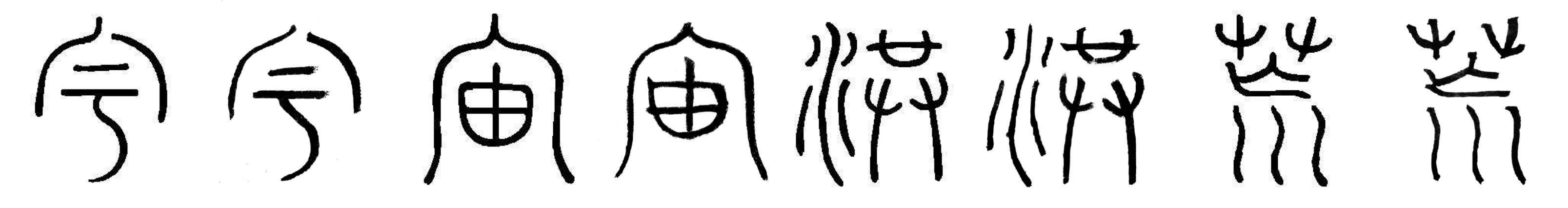}
                \label{fig: ch_result_2}
        \end{minipage}
    }
     \end{minipage}
    \vspace{-4mm}
    \begin{minipage}{\textwidth}
    \subfloat{
        \begin{minipage}{0.48\textwidth}
                \includegraphics[width=1.0\linewidth]{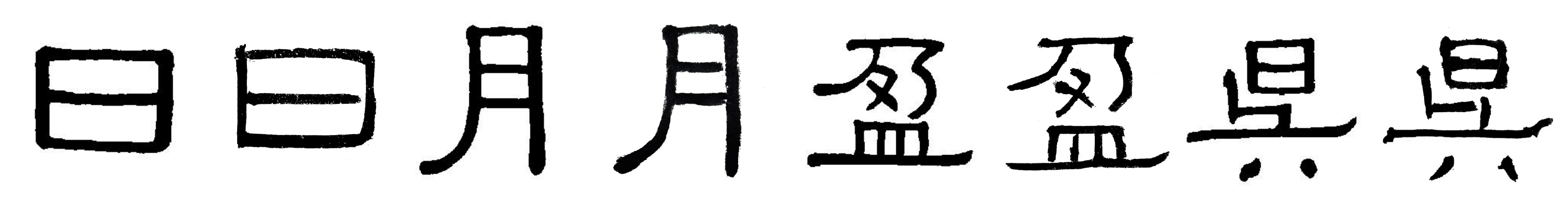}
                \label{fig: ch_result_3}
        \end{minipage}
        \hspace{-3mm}
        \begin{minipage}{0.48\textwidth}
                \includegraphics[width=1.0\linewidth]{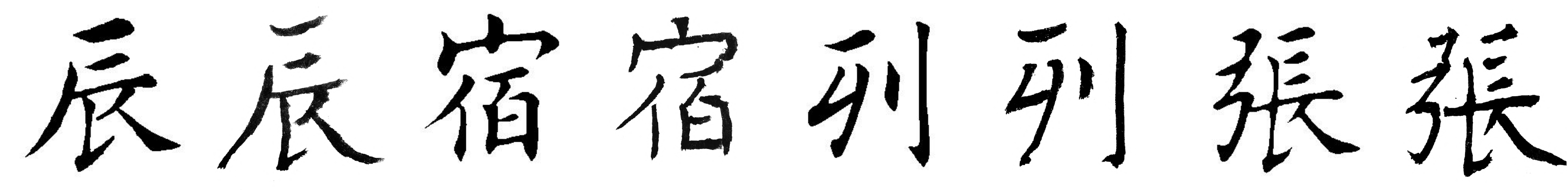}
                \label{fig: ch_result_4}
        \end{minipage}
    }
     \end{minipage}
    \vspace{-4mm}
    \begin{minipage}{\textwidth}
    \subfloat{
        \begin{minipage}{0.48\textwidth}
                \includegraphics[width=1.0\linewidth]{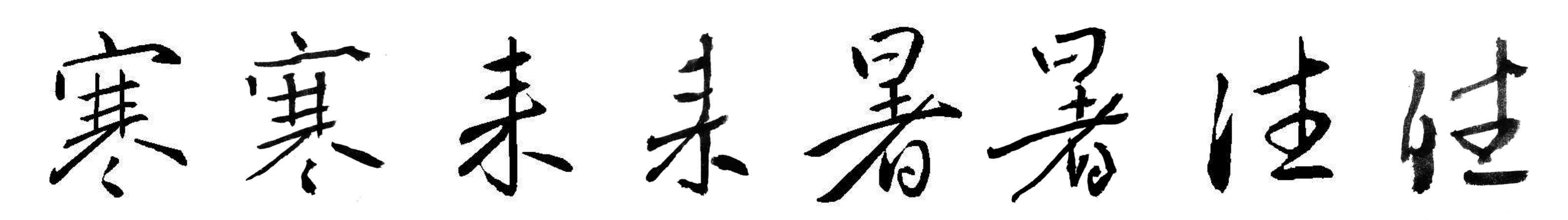}
                \label{fig: ch_result_5}
        \end{minipage}
        \hspace{-2.6mm}
        \begin{minipage}{0.481\textwidth}
                \includegraphics[width=1.0\linewidth]{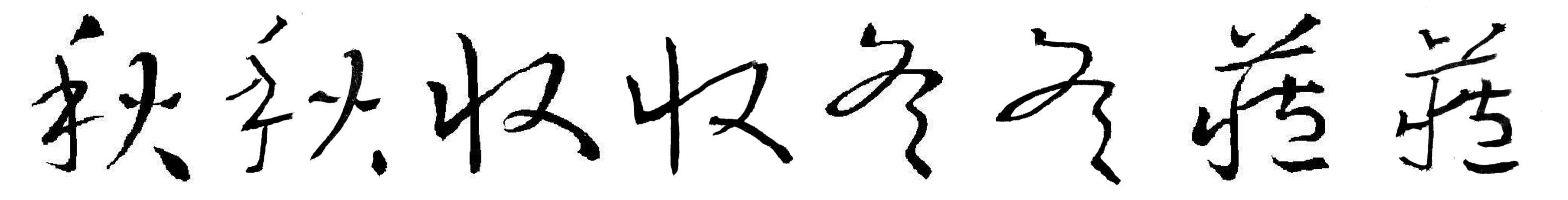}
                \label{fig: ch_result_6}
        \end{minipage}
    }
     \end{minipage}
     \vspace{-1mm}
    \caption{Rewriting various Chinese scripts: the Oracle (left, up); Seal (right, up); Clerical (left, mid); Regular (right, mid); running (left, down) and cursive (right, down). Each pair consists of an input glyph (left) and a robot-generated result (right).} \label{fig: chinese result}
\end{figure*}
\begin{figure*}[t]
\vspace{-4mm}
    \centering
    \begin{minipage}{0.962\textwidth}
    \includegraphics[width=\linewidth]{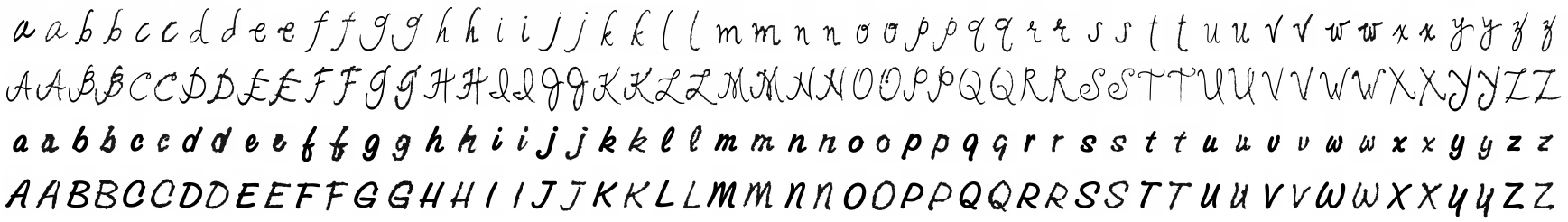}
    \label{fig: eng_result}
    \end{minipage}
    \vspace{-4.5mm}
    \caption{Rewriting two English scripts with fude pen. Each pair includes an input glyph (left) and a robot arm replica (right).} \label{fig: English result}
    \vspace{-4mm}
\end{figure*}
\begin{figure*}[htbp]
    \centering
    \subfloat{
        \hspace{-5mm}
        \begin{minipage}{0.485\textwidth} \includegraphics[width=1.0\linewidth]{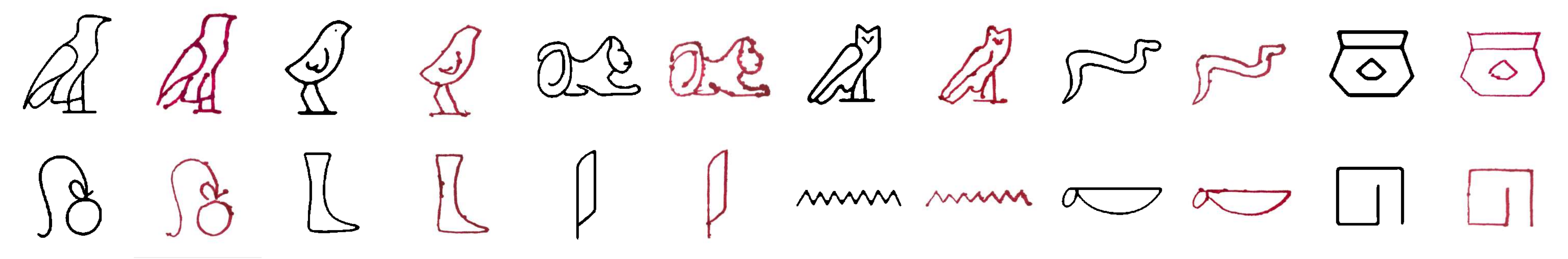}
                \label{fig: egyptian}
        \end{minipage}
        \hspace{-2mm}
        \begin{minipage}{0.485\textwidth}
        \includegraphics[width=1.0\linewidth]{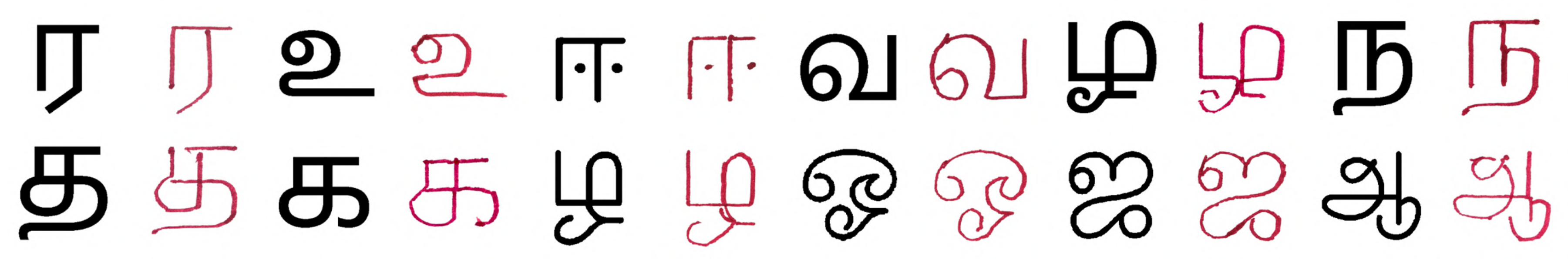}
                \label{fig: tamil}
        \end{minipage}
    }
    \vspace{-6mm}
    \caption{Results on Ancient Egyptian (left) and Tamil (right) characters replicated by Dobot Magician robot arm.} \label{fig: Egyptian and Tamil result}
    \vspace{-4mm}
\end{figure*}
\vspace{-2.3mm}
\subsection{Robotic Demonstration}
\vspace{-1.6mm}
We use a 4-DoF Dobot Magician robot to replicate scripts with physical tools. After optimization, fine control points $(x, y, r)$ are transferred to 3D positions $(x, y, z)$ via calibration for further inverse-kinematics calculation. We adopt the robotic arm with writing tools to leave traces on the paper at varying $z$ heights, measure widths, calculate $r$, and establish the $z$-$r$ connection through linear fitting. We conduct experiments in real-world scenarios, rewriting glyphs in various languages and fonts using calligraphy brushes and fude pens.
\vspace{-4mm}
\subsubsection{Rewriting Chinese Glyphs}
Chinese Calligraphy poses challenges to learners and robots due to its intricate techniques, such as lifting, turning, pressing, and mastering the inelastic soft brush. We select the first 24 characters from the \emph{Thousand Character Classic} with Oracle to Cursive scripts, apply a real calligraphy brush on the robot arm, and set the redrawn size to 8$\times$8 $\text{cm}^2$. Figure~\ref{fig: chinese result} shows our ability to rewrite styles close to the ground truth, with slight distortion in a few characters due to zero adaptation for sim2real.
\vspace{-1.8mm}
\subsubsection{Rewriting English Characters}
We test the fude pen and the flat-tip marker on two English cursive script datasets. The rewriting size of each character is 2$\times$2 $\text{cm}^2$. The robot-written results are shown in Figure~\ref{fig: English result}. The letters redrawn by the robotic arm coincide well with the original glyphs.
\vspace{-1.8mm}
\subsubsection{Test on Other Languages}
We also test our model on unseen characters or hieroglyphs, such as Ancient Egyptian and Tamil. Figure~\ref{fig: strokes} and Figure~\ref{fig: Egyptian and Tamil result} visualize the segmented and rewritten results. In cases when characters are unfamiliar or even like sketches, CalliRewrite still possesses compatible segmentation abilities and can rewrite faithfully.

\section{CONCLUSIONS}
\begin{figure}[t]
    \vspace{-2mm}
    \begin{minipage}{0.5\textwidth}
        \centering
        \includegraphics[width=0.7\linewidth]{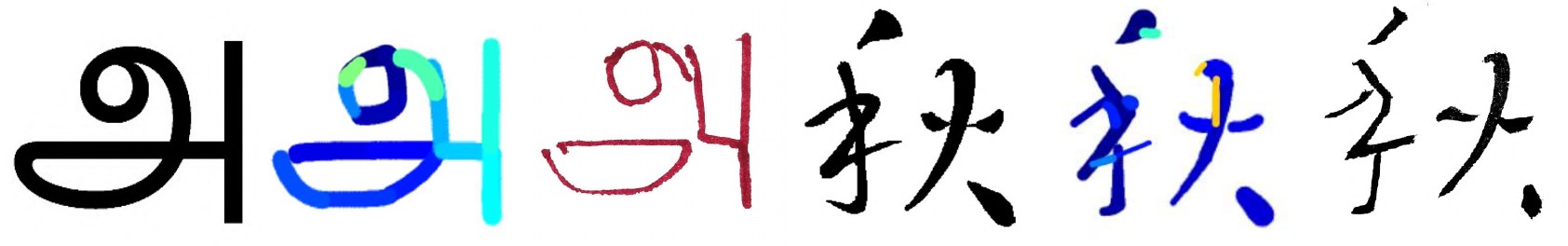}
    \end{minipage}
\vspace{-2mm}
\caption{Failure cases. The right character is harmed by false stroke segmentation, while the left is also caused by error accumulation.}
\label{fig: failures}
\vspace{-6mm}
\end{figure}

\vspace{-1.5mm}
This paper introduced an unsupervised approach for faithfully reproducing diverse characters using different tools, achieving impressive results without any annotation requirements. However, As shown in Figure~\ref{fig: failures}, there also exist some failure cases in our method. This is mainly due to inaccurate stroke segmentation and error accumulation in the robotic arm. These challenges may arise from the model's inherent difficulty in perfectly segmenting typical fonts in unsupervised settings, imprecision in tool modeling, and a lack of real-world training scenarios to bridge the sim2real gap. We leave these challenges as our future work.
\section{ACKNOWLEDGMENT}
\vspace{-1.5mm}
This work was supported by National Natural Science Foundation of China (Grant No.: 62372015), Center For Chinese Font Design and Research, and Key Laboratory of Intelligent Press Media Technology.






\newpage
\bibliographystyle{IEEEtran}
\newpage
\bibliography{IEEEabrv, reference}

\end{document}